\documentclass[journal]{IEEEtran}
\usepackage{etoolbox}
\makeatletter
\patchcmd{\@makecaption}
  {\scshape}
  {}
  {}
  {}
\makeatother

\usepackage{graphicx}
\usepackage{amsmath,amssymb} 
\usepackage{color}
\usepackage{changes}
\usepackage{booktabs} 
\usepackage{float}
\usepackage{todonotes}
\usepackage{colortbl}

\usepackage{hyperref}
\hypersetup{
  linkcolor=blue;
  citecolor=green;
}
\usepackage{cite}
\ifCLASSINFOpdf
\else
\fi
\usepackage{amsmath}
\usepackage{url}

\begin{document}

\title{Coarse-to-fine Semantic Segmentation from Image-level Labels}

\author{Longlong~Jing*,
        Yucheng~Chen*,
        and Yingli~Tian$^\dag$, 
        Fellow, \textit{IEEE}

\thanks{$^*$Equal contributions; $^\dag$Corresponding author.}

\thanks{L. Jing is with the Department of Computer Science, The Graduate Center, The City University of New York, NY, 10016. E-mail:ljing@gradcenter.cuny.edu.}

\thanks{Y. Chen is with School of Electronics and Information, the Northwestern Polytechnical University, Shaanxi Xi'an, China, 710129. This work is done when he is visiting the City College of New York. E-mail:chenyucheng@mail.nwpu.edu.cn.}

\thanks{Y. Tian is with the Department of Electrical Engineering, The City College, and the Department of Computer Science, The Graduate Center, The City University of New York, NY, 10031. E-mail: ytian@ccny.cuny.edu.}

\thanks{This material is based upon work supported by the National Science Foundation under award numbers IIS-1400802 and EFRI-1137172. Manuscript received December 16, 2018.}}

\maketitle

\begin{abstract}
Deep neural network-based semantic segmentation generally requires large-scale cost extensive annotations for training to obtain better performance. To avoid pixel-wise segmentation annotations which are needed for most methods, recently some researchers attempted to use object-level labels (e.g. bounding boxes) or image-level labels (e.g. image categories). In this paper, we propose a novel recursive coarse-to-fine semantic segmentation framework based on only image-level category labels. For each image, an initial coarse mask is first generated by a convolutional neural network-based unsupervised foreground segmentation model and then is enhanced by a graph model. The enhanced coarse mask is fed to a fully convolutional neural network to be recursively refined. Unlike existing image-level label-based semantic segmentation methods which require to label all categories for images contain multiple types of objects, our framework only needs one label for each image and can handle images contains multi-category objects. With only trained on ImageNet, our framework achieves comparable performance on PASCAL VOC dataset as other image-level label-based state-of-the-arts of semantic segmentation. Furthermore, our framework can be easily extended to foreground object segmentation task and achieves comparable performance with the state-of-the-art supervised methods on the Internet Object dataset.
\end{abstract}

\begin{IEEEkeywords}

Weakly Supervised Learning, Semantic Segmentation, Foreground Object Segmentation, Convolutional Neural Network, Deep Learning

\end{IEEEkeywords}

%
\IEEEpeerreviewmaketitle

\section{Introduction}
\IEEEPARstart{S}{emantic} segmentation, the task of assigning semantic labels to each pixel in images, is of great importance in many computer vision applications such as autonomous driving, human-machine interaction, and image search engines. The community has recently made promising progress by applying Convolutional Neural Network (CNN) due to its powerful ability to learn image representations. Various networks such as FCN \cite{FCN}, DeepLab \cite{DeepLab}, PSPNet \cite{PSP}, SegNet \cite{SegNet} and datasets such as PASCAL VOC \cite{VOC}, CityScape \cite{Cityscape}, CamVid \cite{CamVid}, ADE20K \cite{ADE20K} have been proposed for semantic segmentation. 

\begin{figure*}[!ht]
\centering
\includegraphics[width=0.95\textwidth]{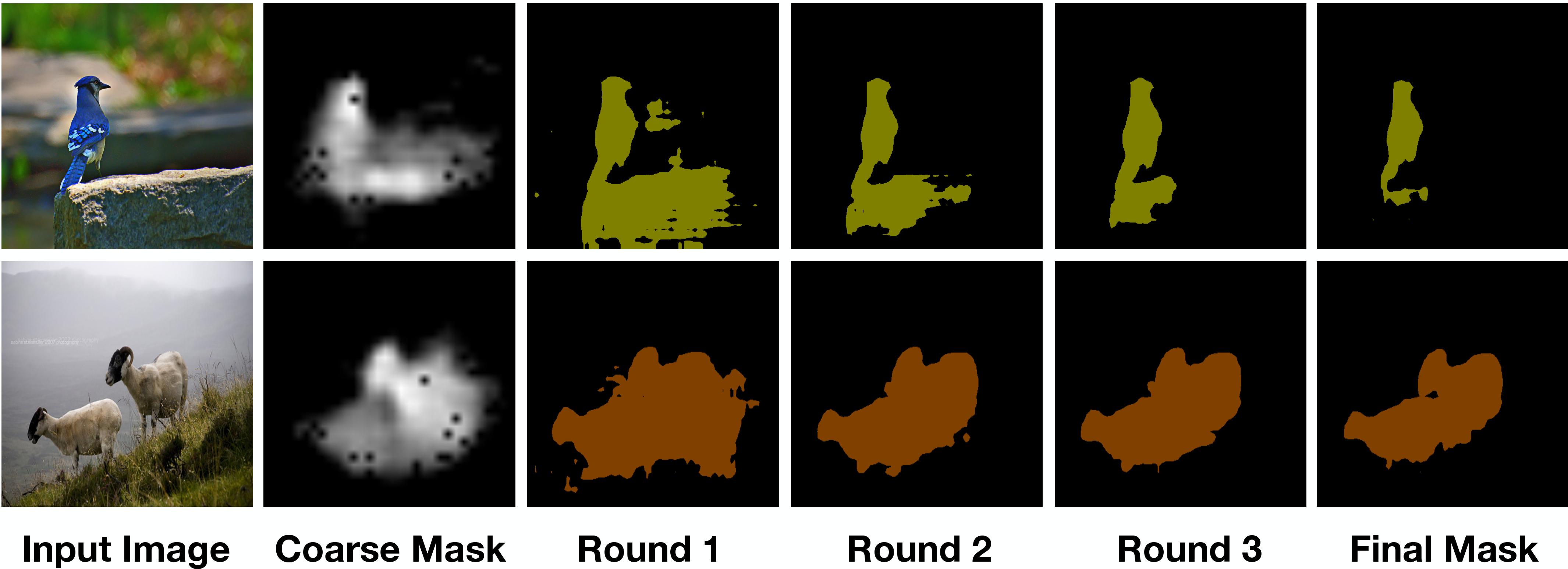}
\caption{Two sets of images and their corresponding refinement masks of different training rounds. Trained with simple image category labels and coarse masks, our framework can finally predict fine segmentation masks for the input images.}
\label{fig:demo_img}
\vspace{-10pt}
\end{figure*}

The performance of deep neural network (DNN) greatly depends on the capability of the network and the amount of training data. Different kinds of network architectures were developed to increase the capacity of the models, and larger and larger datasets were keeping proposed these days. However, even though several datasets have been annotated for semantic segmentation, the amount of training data for semantic segmentation is still relatively small compared to the datasets for other tasks such as ImageNet \cite{ImageNet} and YouTube Bounding Box \cite{YoutubeBB}. Semantic segmentation generally requires pixel-wise semantic annotation which makes the cost of annotation process time-consuming and expensive.  

To mitigate the limitation of the annotations, weakly supervised and semi-supervised semantic segmentation methods were proposed \cite{semi_gan}, \cite{web_videos}, \cite{simple_does}, \cite{region_mining}, \cite{scribble}. By utilizing annotations that cheaper than pixel-wise annotation such as object-level labels including bounding boxes, scribbles, spots, or image-level labels to train the semantic segmentation models greatly reduce the cost of data annotation. Furthermore, these annotations can be easily obtained to produce large weakly supervised datasets. Trained with the weak labels, these models achieve promising performance and the gap between the weakly supervised and supervised methods in performance are getting smaller. However, these methods still need cumbersome labors such as accurate bounding boxes \cite{simple_does}, \cite{BoxSup}, \cite{WSSL} and scribbles \cite{scribble}, \cite{one_point}. For example, in the model trained with the bounding boxes, all the interested category of objects must be annotated with accurate bounding boxes.

In this paper, we propose a novel semantic segmentation framework to be trained with images directly retrieved from a subset of the ImageNet dataset while only the image category labels are available. The cost of obtaining image-level labels is much lower than object-level annotations such as bounding boxes, spots, and scribbles. Unlike other weakly supervised methods, it is worth mentioning that the category information of the images in our training dataset is very simple and inaccurate. In others' work such as \cite{semi_gan}, \cite{web_videos}, \cite{region_mining} \cite{AF_MCG}, one image usually has multiple labels of the interested object categories appeared in the image, however, one image has been labeled only one category in ImageNet even when the image contains objects of multiple categories. Even though trained on the dataset with simple and inaccurate labels, our model can generate semantic segmentation masks for images containing multi-category objects. 

Our goal is to train the segmentation network only with image-level labels. Specifically, we train the model with images directly from the ImageNet dataset \cite{ImageNet}, which is collected and labeled for image classification task. Some example images and the corresponding coarse-to-fine masks generated by our framework are illustrated in Fig.~\ref{fig:demo_img}. Firstly, the student network is employed to generate a coarse mask for each image which was proposed for unsupervised learning foreground segmentation in images by Croitoru \textit{et al.} \cite{forground_seg}. Since the coarse masks generally are very rough and have many holes and their locations are inconsistent with the object locations, a graph model is employed to enhance the coarse masks. Then, the enhanced masks, the input images, and the category labels of the images are used to recursively train the semantic segmentation network which is a fully convolutional network. Trained with only category information, the network can finally generate pixel-wise semantic masks for the input images.


In summary, our main contributions are: 
\begin{itemize}
  
  \item{We propose a new weakly supervised framework for semantic segmentation only depends on image category level annotations.} 

  \item{The proposed framework recursively refines the coarse masks to fine, while the coarse masks are generated by an unsupervised foreground segmentation method and enhanced by a graph model.}

  \item{Trained with images that each has only one category label, the proposed framework can generate final pixel-wise semantic fine masks for images containing multiple category objects.} 

   \item{The proposed framework can be generalized cross datasets. With only trained on a subset of the ImageNet dataset, and it achieves comparable performance on PASCAL VOC dataset as other image-level label-based state-of-the-arts.}
   
   \item {The proposed framework can be easily extended to image foreground object segmentation and it achieves comparable performance with the state-of-the-art supervised methods on the Internet Object dataset.}
 \end{itemize}

\section{Related Work}

Recently, many semantic segmentation methods have been proposed. Based on the level of annotations used, these methods fall into three categories: fully supervised pixel-wise annotation-based methods that trained with pixel-wise labels annotated by human labors \cite{FCN}, \cite{DeepLab}, \cite{PSP}, \cite{SegNet}; weakly supervised object-level annotation-based methods that trained with object-level annotations such as bounding boxes, spots, and scribbles \cite{simple_does}, \cite{scribble}, \cite{WSSL}, \cite{one_point}; and weakly supervised image-level annotation-based methods that trained with image category labels \cite{semi_gan}, \cite{web_videos}, \cite{region_mining}. Trained with accurate pixel-labels, fully supervised pixel-wise annotation-based methods have the best performance.

\textbf{Fully supervised pixel-wise annotation-based methods:} Long \textit{et al.} \cite{FCN} made the first attempt to apply fully convolutional network (FCN) in semantic segmentation and achieved the milestone break. Badrinarayanan \textit{et al.} proposed a symmetric auto-encoder architecture by utilizing the convolution and deconvolution layers \cite{SegNet}. Chen \textit{et al.} \cite{DeepLab} employed the atrous convolution, atrous spatial pyramid pooling, and fully-connected Conditional Random Field (CRF) in semantic segmentation, which was widely used in other networks later. Zhao \textit{et al.} \cite{PSP} proposed to employ the pyramid pooling model to aggregate the context information of different regions in an image and achieved the state-of-the-art performance on various semantic segmentation datasets.

\begin{figure*}[!ht]
\centering
\includegraphics[width=\textwidth]{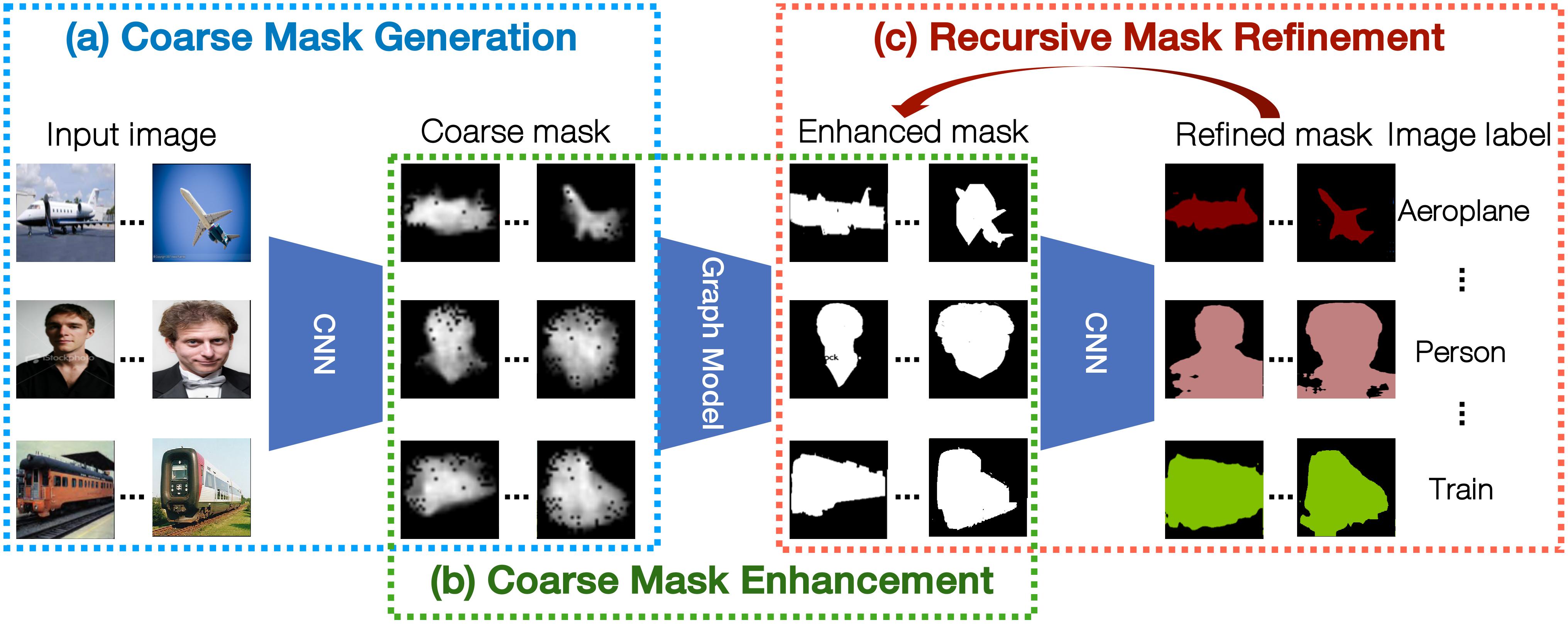}
\caption{ The main components of the proposed method: (a) coarse mask generation; (b) coarse mask enhancement; and (c) recursive mask refinement.  }
\label{fig:pipeline}
\end{figure*}

\textbf{Weakly supervised object-level annotation-based methods:} Object bounding boxes, as a relatively cheaper yet quite accurate annotation, are used to train weakly supervised semantic segmentation models. In this way, bounding boxes annotated for other tasks such as object detection can be directly used to train segmentation models. Papandreou \textit{et al.} \cite{WSSL} developed an Expectation-Maximization (EM) method for semantic image segmentation model trained on the bounding boxes annotations and obtained competitive results. Dai \textit{et al.} \cite{BoxSup} proposed a CNN model trained on bounding boxes of automatically generated region proposals. Khoreva \textit{et al.} \cite{simple_does} proposed to train with bounding boxes for semantic and instance segmentation. With the relatively accurate annotations and powerful model, they achieved the state-of-the-art results in weakly supervised semantic segmentation. Lin \textit{et al.} proposed to train a network with scribbles which are the minimum object-level annotations \cite{scribble}.

\textbf{Weakly supervised image-level annotation-based methods:} Wei \textit{et al.} \cite{region_mining} proposed to train the model with image-level labels by transferring the image classification models into segmentation model via adversarial training. Hong \textit{et al.} \cite{web_videos} proposed to utilize videos collected by web engines, along with the weakly annotated images to train the models. Taking the advantages of Generative Adversarial Networks (GANs), fake images generated by GAN along with some real images with image-level labels are used to train a segmentation model by Souly \textit{et al.} \cite{semi_gan}. Hong \textit{et al.} \cite{TransferNet} proposed to train a segmentation network with some auxiliary segmentation annotations for different categories and image-level class labels. Kolesnikov and Lampert \cite{SEC} proposed a new loss function for weakly supervised semantic segmentation by constraining the segmentation to coincide with object boundaries. Qi \textit{et al.} proposed to implicitly utilize the stronger supervision to guide the weakly segmentation model \cite{AF_MCG}. Wei \textit{et al.} \cite{wei2018revisiting} proposed to employ dilated convolution to generate reliable object localization maps. Zhang \textit{et al.} \cite{zhang2018decoupled} proposed a decoupled spatial neural attention network to generate pseudo-annotations by localizing the discriminative parts of the object region.


\textbf{Foreground Object Segmentation:} There are three main strategies for generic foreground object segmentation: joint segmentation-based methods which use the prior knowledge as the supervision \cite{InternetObjectdataset}, \cite{joulin2010discriminative}, \cite{joulin2012multi}, \cite{kim2011distributed}, \cite{chen2014enriching}, \cite{activesegmentation}, saliency prediction-based methods which identify regions likely to capture human attention \cite{DeepMC}, \cite{deepsaliency}, \cite{MarkovChain}, \cite{booleanmap}, and object proposal-based methods which localize all the objects in images \cite{MCG}, \cite{DeepMask}, \cite{SalObj}. Jiang \textit{et al.} proposed to formulate saliency detection via absorbing Markov chain on an image graph model \cite{MarkovChain}. Zhang and Sclaroff proposed a boolean map-based model to predict the saliency. Each image is characterized by a set of binary images, and then saliency maps are generated by analyzing the topological structure of these boolean maps \cite{booleanmap}. Since low-level cues or priors do not produce good enough saliency detection, Zhao \textit{et al.} employed multi-context deep learning framework to model saliency of objects in images by utilizing multi-context features\cite{DeepMC}. Arbel{\'a}ez \textit{et al.} proposed a Multiscale Combinatorial Grouping (MCG) for bottom-up hierarchical image segmentation and object candidate generation \cite{MCG}. Pinheiro \textit{et al.} proposed to train a discriminative convolutional neural network with multiple objectives, while one of them is to generate a class-agnostic segmentation mask \cite{DeepMask}. Jain \textit{et al.} proposed to train a fully convolutional neural network, which was originally designed for semantic segmentation, for the foreground object segmentation \cite{pixelobjectness}.

Different from other image-level based methods that need multi-category information for each image, we propose a new coarse-to-fine framework for semantic segmentation by using images with only one category label and achieves comparable performance with the state-of-the-art weakly supervised semantic segmentation methods. Furthermore, our framework can be easily extended to foreground segmentation task and achieves comparable performance with the state-of-the-art supervised methods on the Internet Object dataset \cite{InternetObjectdataset}. 

\section{The Proposed Approach}

\subsection{Overview}
As shown in Fig.~\ref{fig:pipeline}, our framework contains three main components: coarse mask generation, coarse mask enhancement, and recursive mask refinement. Firstly, a trained 8-layer CNN is employed to generate the initial coarse masks for images. Secondly, a graph-based model is employed to enhance the quality of the initial coarse masks based on the object prior. Finally, these enhanced masks together with the input images and their category labels are used to recursively train a fully convolution network designed for semantic segmentation.

\begin{figure*}[!ht]
\centering
\includegraphics[width=\textwidth]{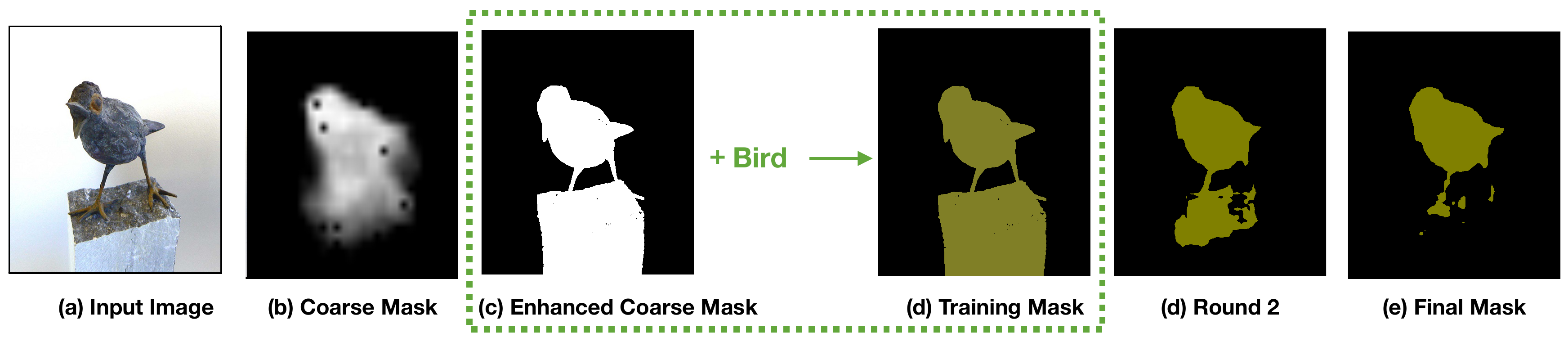}
\caption{The process of semantic mask generation. The training mask is obtained by assigning the image category to the foreground pixels in the enhanced masks.}
\label{fig:grabcut_demo}
\end{figure*}

\subsection{Coarse Mask Generation}

The core intuition behind this step is to generate coarse masks without any class labels. Many methods can generate the coarse masks such as VideoPCA \cite{VideoPCA}, Non-Local Consensus Voting (NLC) \cite{NLC}, Unsupervised Foreground Segmentation(UFS) \cite{forground_seg}, and Unsupervised Object Segmentation (UOS) \cite{VideoPCA_new}. These methods can segment moving objects in videos or generate saliency maps for images. 

Different from other methods that perform unsupervised object discovery in videos or in collections of images at testing time \cite{VideoPCA}, \cite{NLC}, \cite{VideoPCA_new}, the framework in \cite{forground_seg} is a CNN-based network that trained with millions of unlabeled images and achieves the state-of-the-art in unsupervised object segmentation. Moreover, the student network, an 8-layer CNN trained on large scale video frames, in \cite{forground_seg} is two orders of magnitude faster than other previous methods at testing. The coarse masks can be obtained by applying a standard feed-forward processing along the network. Therefore, the student network is employed to generate the coarse masks in our framework.

However, the generated coarse masks are very noisy and inaccurate. As shown in Fig.~\ref{fig:demo_img}, usually there are many holes and the locations of the masks are inconsistent with the locations of the objects. The quality of masks is essential to the performance of the semantic segmentation. Inspired by \cite{simple_does}, \cite{scribble}, a graph model is employed to enhance the masks to train the semantic segmentation network. 

\subsection{Coarse Mask Enhancement}

The semantic segmentation network would have an inferior performance by directly train on noisy and inaccurate coarse masks. Therefore, a mask enhancement is conducted before the recursive training of semantic segmentation network. 

Following \cite{simple_does}, \cite{scribble},  \cite{forground_seg}, GrabCut \cite{GrabCut} is employed as an unsupervised mask enhancement technique to improve the quality of the initial coarse masks. GrabCut is an efficient interactive foreground/background segmentation method based on graph cuts \cite{GrabCut}. The model employs a Gaussian Mixture Model to estimate the color distribution of the foreground objects and that of the background. These distributions are then used to construct a Markov Random Field over the pixel labels (i.e. foreground or background). A graph cut-based optimization method is run to minimizing the energy function that prefers connected regions having the same label. By repeating the two-step procedure until it converges, the enhanced coarse masks are obtained.

Some example images and their corresponding coarse masks, the enhanced masks are shown in Fig.~\ref{fig:pipeline}. The location and shape of the enhanced masks are more accurate and compact than the coarse masks. These enhanced masks are used to recursively train the semantic segmentation network. 

\subsection{Recursive Mask Refinement}

So far for each image, both the generated coarse mask and enhanced mask are obtained as foreground by unsupervised learning without semantic category labels. We propose a recursive semantic segmentation network to obtain the pixel-wise semantic segmentation mask by combining the image category label with the enhanced coarse mask as the initial semantic labels for training. Our semantic segmentation network is trained on a subset of the ImageNet dataset. Since ImageNet dataset is designed for image classification problem with the main object usually occupies a large space in the image, the enhanced coarse masks can cover most part of the main object. Based on this assumption, the category label of each image is assigned to all the pixels belong to the foreground object, and the category for the rest of the pixels is set to the background. This process is demonstrated in Fig.~\ref{fig:grabcut_demo}. These generated enhanced masks are taken as the initial input in the next training round of our recursive semantic segmentation network.

We choose DeepLab \cite{DeepLab} as the semantic segmentation network due to its practical merit and effectiveness. By using the atrous convolution to increase the receptive field of neurons, employing the atrous spatial pyramid pooling (ASPP) to consider the objects at multiple scales, and using the Fully Connected CRF to improve the localization performance of object boundaries, Deeplab achieved the state-of-the-art in semantic segmentation benchmarks. At the end of the first training round, we obtain a semantic segmentation model that can be applied to any image to predict the semantic mask. Since the quality of the enhanced masks is low and the model can hardly reach its capacity just in one training round, we propose to recursively train the network to continue to refine the semantic masks by taking the output masks from the last training round as the input masks of the current training round. This process repeats for several iterations until the network converges. Some example images and their corresponding masks after each training round can be found in Fig.~\ref{fig:refinement_visualize}. We only keep the pixels have the same category as the image as the new mask, the category of the rest of the pixels is set to background. Then GrabCut is applied on this new mask to enhance it based on the object prior. Finally, these enhanced masks are used as the semantic labels to continue to train the network. 

When the training finished, we obtain the segmentation network that can segment the interested category out for any given image. In the training phase of the semantic segmentation model, the only annotation needed is the image-level labels which designed for image classification. Also, no human-made labels are needed to train the student network in \cite{forground_seg}.

\subsection{Extend the Proposed Framework to Foreground Segmentation}

In order to demonstrate the generalization of the proposed framework, we further extend it for foreground segmentation task. Compared to the semantic segmentation which needs to recognize the category of each pixel, the general foreground object segmentation only needs to identify whether the pixel belongs to foreground objects or not. Therefore,  by only replacing the network in the "recursive mask refinement" step with the network that designed for foreground object segmentation, our framework can be applied to foreground segmentation task. Due to its powerful ability to fuse multi-scale features, we choose the Feature Pyramid Network \cite{FPN} as the network for foreground object segmentation task.

\section{Experiments}

To evaluate our proposed framework, we conduct several experiments including the impact of quality of masks, the effect of the number of training round, and compared with others work. All our experiments are trained on the subset of the ImageNet dataset with only category labels and evaluated on the PASCAL VOC dataset with the same set of parameters.

\begin{figure*}[!h]
\centering
\includegraphics[width=0.8\textwidth]{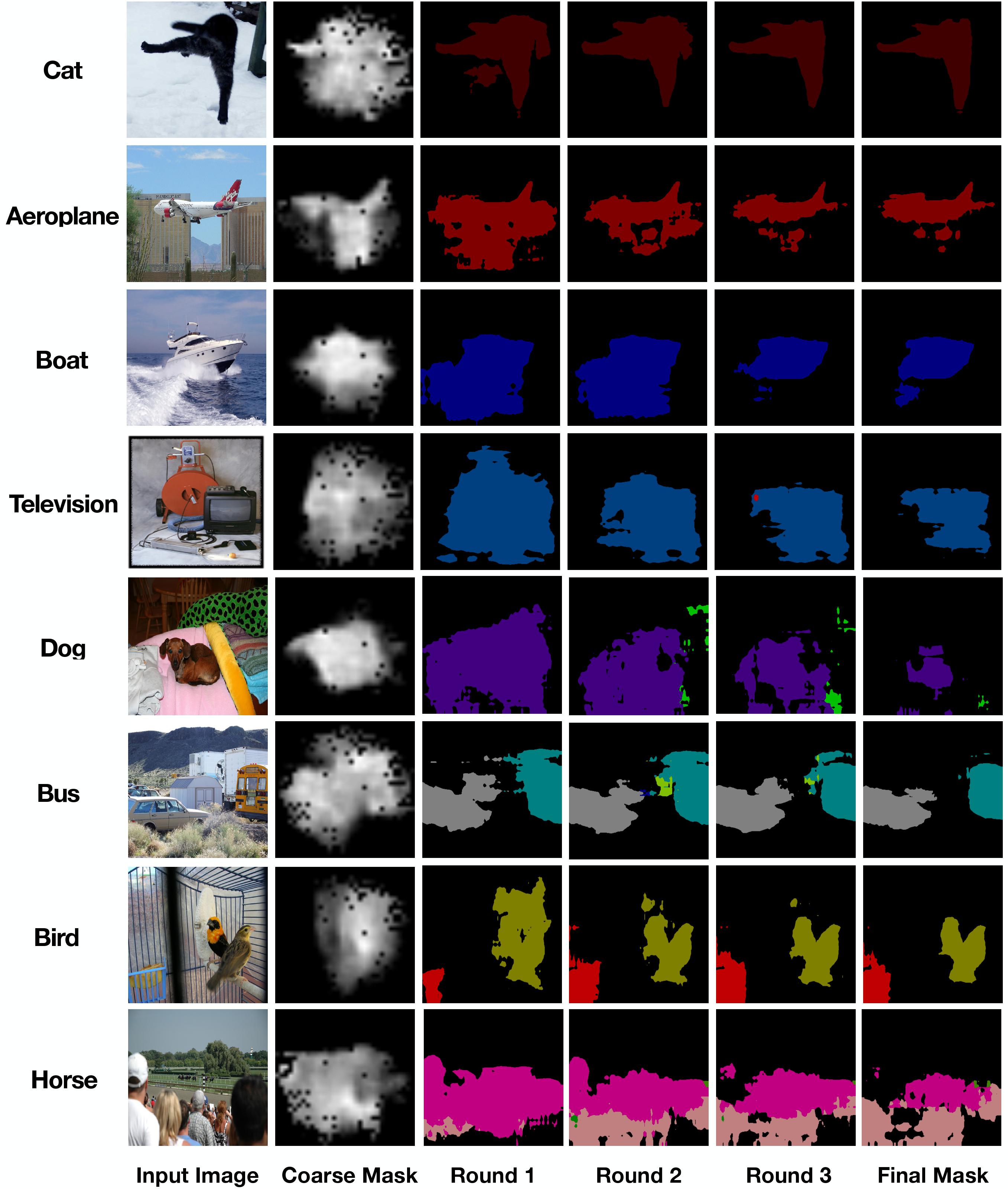}
\caption{Example semantic segmentation results from coarse to fine. Our model can obtain the semantic masks for multiple category objects appear in the same image as shown in the last three rows. Different colors within one image mask represent different category objects. }
\label{fig:refinement_visualize}
\end{figure*}
\vspace{-10pt}

\subsection{Datasets}

\textbf{ImageNet:} ImageNet is an image collection organized according to the WordNet hierarchy, described by multiple words or word phrase, is called a  ``synonym set'' or ``synset'', each expressing one concept. There are more than $100,000$ synsets in WordNet. {The ImageNet aims to provide an average of $1,000$ images to illustrate each synset, which is quality-controlled and human annotated with image-level labels.} We select a subset of images from ImageNet belonging to $20$ categories which are same as that in VOC dataset without including the background category. The list of the synsets used in the experiments can be found in the supplementary material.

\textbf{VOC:} The semantic segmentation network is evaluated on the PASCAL VOC $2012$ segmentation benchmark dataset \cite{VOC} containing $21$ object categories including the background. The dataset is split into three subsets: $1,464$ images for training, $1,449$ images for validation and $1,456$ images for testing. For a fair comparison with other weakly supervised  image-level based state-of-the-art semantic segmentation methods \cite{region_mining}, \cite{BoxSup}, \cite{AF_MCG}, \cite{SEC}, \cite{BFBP}, we use the same validation and test datasets to as others to obtain the segmentation results. Since the ground truth masks for the VOC testing dataset is not released, the testing accuracy is obtained by submitting the predicted results to the PASCAL VOC evaluation server.

\textbf{MIT Object Discovery Dataset:} The foreground object segmentation network is evaluated on the MIT Object Discovery Dataset which contains $2,488$ images belonging to three categories of foreground objects: Airplanes, Cars, and Horses. The images in this dataset were collected from Internet search, and each image is annotated with pixel-level labels for the evaluation purpose. This dataset is most widely used for evaluating weakly supervised foreground segmentation methods. For a fair comparison with other weakly supervised methods \cite{pixelobjectness}, the performance on the same test set and a subtest set of this dataset are reported and compared with others.


\begin{figure}[!h]
\centering
\includegraphics[width=0.5\textwidth]{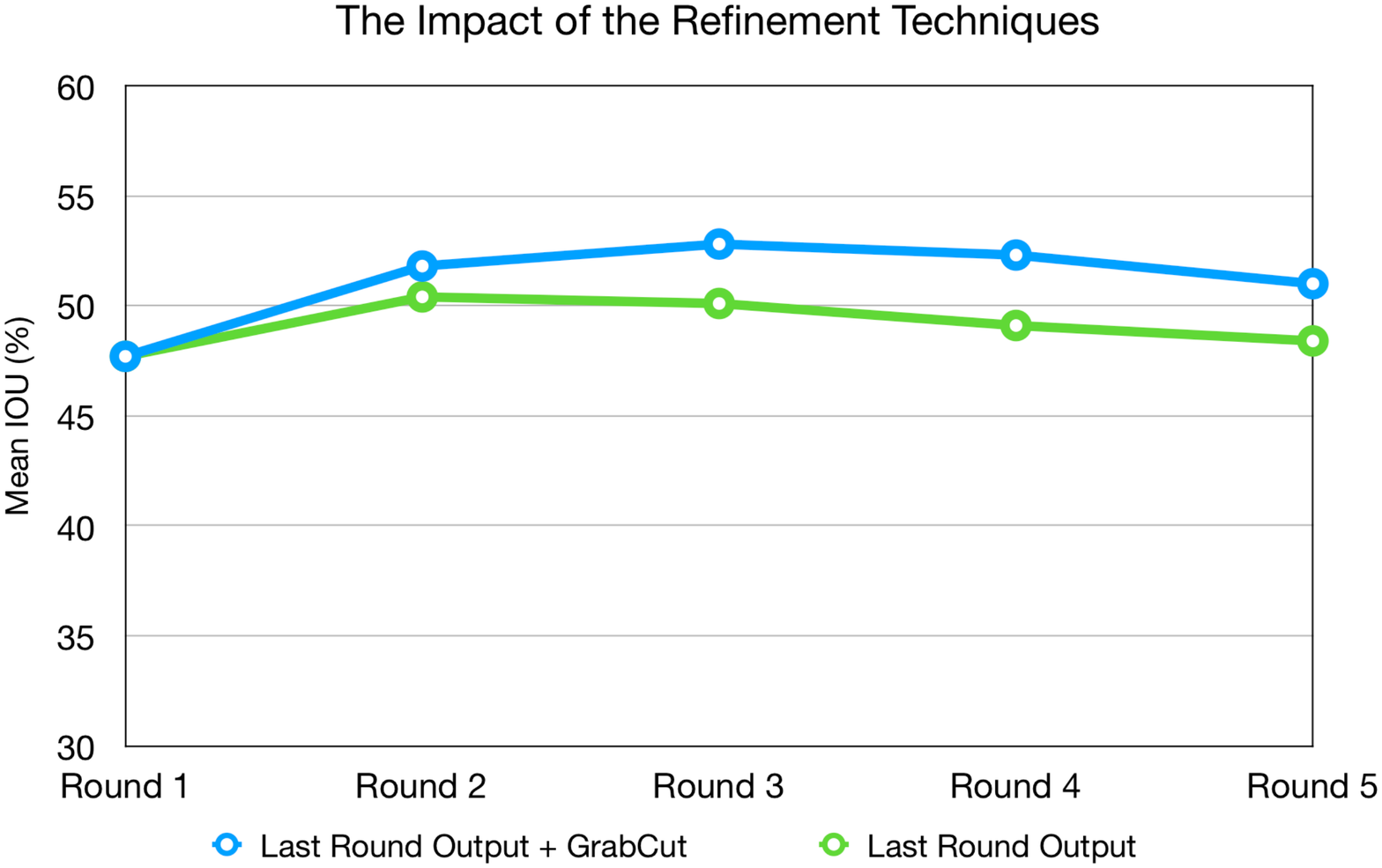}
\vspace{-30pt}
\caption{The performance of semantic segmentation with and without GrabCut on the validation split of PASCAL VOC dataset. }
\label{fig:voc_performance}
\end{figure}

\subsection{Training Details for Recursive Mask Refinement}

We use mini-batch Stochastic Gradient Descent (SGD) with a batch size of $12$ images. The learning rate is set to $0.00025$, momentum to $0.9$, and weight decay to $0.0005$. This network is trained with images from our collected subset of the ImageNet. The refined masks of each image are generated by our recursive semantic segmentation network followed by GrabCut \cite{GrabCut} process. Training is completed for $5$ training rounds and the masks are updated at the end of each training round. During training, random crop and resize are applied for data augmentation. Specifically, each image is resized to $321 \times 321$, then random cropped to the size of $256 \times 256$. There are around $9,000$ images in our training dataset belonging to $20$ object categories. We use the non-foreground regions within these images as the background in the training phase.

\subsection{Evaluation Metrics}

Following others \cite{semi_gan}, \cite{web_videos}  \cite{simple_does}, \cite{BoxSup}, \cite{WSSL}, Intersection over Union (IOU) which is averaged across $21$ categories ($20$ for objects and one for background) are computed to evaluate the performance for the semantic segmentation. We conduct experiments on the validation split to guide our experiment design. Final results are reported and compared with other methods on the test split of PASCAL VOC dataset. For the foreground object segmentation task, the IOU of the predicted binary mask of the foreground object and the ground truth mask is calculated and compared with the state-of-the-art methods.

\subsection{Semantic Segmentation Results}

\subsubsection {Impact of the mask quality.} To evaluate the impact of the quality of the mask to the performance, we compare the mean IOU of DeepLabs trained with four kinds of masks: coarse masks, enhanced masks, bounding boxes of enhanced masks, and refined masks at the end of the first training round. During the training, these masks are taken as the semantic labeled masks. These experiments are only for evaluating the impact of the quality of masks on the performance, we do not recursively train the models in these experiments.

As shown in Table~\ref{tab:cmp_masks_per}, among all the methods, the network trained with coarse masks has the worst performance due to the very low mask quality (many holes and inaccurate). By applying the enhancement with GrabCut, the IOU improves $15.2$\% which comes from the quality improvement. This demonstrates the importance of the quality of masks and the effectiveness of enhancement. The mean IOU of bounding box method is $2.7$\% higher than that of using the coarse masks. This is probably because the enhanced masks are more compact than the coarse masks, and the locations are aligned more closely to the objects. The performance of refined masks is $2.7$\% higher than that of the network trained on the enhanced masks. This validates our idea that the network can refine the masks from coarse to fine. Based on this observation, we recursively train the semantic segmentation network to refine the masks.

\begin{table}
\begin{center}
\caption{The performance of semantic segmentation networks trained with four kinds of masks evaluated in the validation split of PASCAL VOC dataset in mean IOU.} 
\label{tab:cmp_masks_per}
\begin{tabular}{ll}
\hline\noalign{\smallskip}
Training Mask Type & IOU (\%)\\
\noalign{\smallskip}
\hline
\noalign{\smallskip}
Coarse Masks   &$32.5$ \\
Bounding Boxes  & $35.2$  \\
Enhanced Masks  &  $47.7$\\
Refined Masks & $50.4$\\ 
\hline
\end{tabular}
\vspace{-18pt}
\end{center}
\end{table}

\subsubsection {Effectiveness of recursive refinement.} To evaluate the effectiveness of the recursive refinement, we recursively train the models with the two kinds of masks respectively: with and without GrabCut post-processing. At each training round, the masks are updated with the processed output of the network from the last round. Between each training round, the masks are improved by a post-processing with three strategies: a) If less than $1$\% pixels or more than $80$\% pixels are foreground, then this image would not be used to train the network in next training round. b) Since one image has only one label, if network predicts multiple categories for an image, only the pixels belonging to the original category would be valid, all other pixels would be set as background. c) For the recursive training, the GrabCut is applied on the predicted masks by DeepLab to refine the masks. At the end of each training round, the network predicts the masks of all the training images, then the three strategies are applied on all the predicted masks to update the masks. 

\begin{table}[]
    \centering
    \caption{Comparison of weakly-supervised semantic segmentation methods on PASCAL VOC 2012 \emph{test} and \emph{validation} dataset. (* indicates the methods implicitly use pixel-level or other supervisions.)} 
    \label{tab:val_comp}
    \begin{tabular}{lcc}
        \toprule
        \hline
        Methods & mIoU (val) & mIoU (test)   \\
        \midrule
        \multicolumn{2}{l}{Supervision: Bounding Box}  \\
        WSSL (ICCV 2015)~\cite{WSSL}  & $60.6$ & $62.2$ \\
        BoxSup (ICCV 2015)~\cite{BoxSup}  & $62.0$ & $64.2$\\        
        SDI (CVPR 2017)~\cite{simple_does}  &$65.7$ &$67.5$\\
        \midrule
        \multicolumn{2}{l}{Supervision: Scribbles}  \\
        Scribblesup (CVPR 2016)~\cite{scribble}  & $63.1$ & --- \\
        \midrule
        \multicolumn{2}{l}{Supervision: Spot}  \\
        1 Point (ECCV 2016)~\cite{one_point}  & $46.1$ & ---\\
        Scribblesup (CVPR 2016)~\cite{scribble} & $51.6$ & --- \\
        \midrule
        \multicolumn{2}{l}{Supervision: Image-level Labels}  \\    
        MIL-seg* (CVPR 2015)~\cite{MILSEG}  & $42.0$ & $40.6$\\
        TransferNet* (CVPR 2016)~\cite{TransferNet} & $52.1$ & $51.2$\\
        AF-MCG* (ECCV 2016)~\cite{AF_MCG}  &  $54.3$ &  $55.5$\\
        WSSS* (CVPR 2017)~\cite{web_videos}  &  $58.1$ &  $58.7$\\
        \midrule
        MIL-FCN (ICLR 2015)~\cite{MIL_FCN}  & $25.7$ & $24.9$\\
        CCNN (ICCV 2015)~\cite{CCCN}  & $35.3$ & $35.6$\\
        MIL-sppxl (CVPR 2015)~\cite{MILSEG}   & $36.6$ & $35.8$\\
        MIL-bb (CVPR 2015)~\cite{MILSEG} & $37.8$ & $37.0$\\
        EM-Adapt (ICCV 2015)~\cite{WSSL}  & $38.2$ & $39.6$\\
        DCSM (ECCV 2016)~\cite{DCSM}  & $44.1$ & $45.1$\\
        BFBP (ECCV 2016)~\cite{BFBP}  & $46.6$ & $48.0$\\
        STC (PAMI 2017)~\cite{STC}    &$49.8$    & $51.2$\\
        SEC (ECCV 2016)~\cite{SEC}  & $50.7$ & $51.7$\\         
        AF-SS (ECCV 2016)~\cite{AF_MCG}  &  $52.6$ &  $52.7$\\
        WebSeg (CVPR 2017)~\cite{WeblySeg} &  $53.4$ &  $55.3$\\
        AE-PSL (CVPR 2017)~\cite{region_mining}  & $55.0$  & $55.7$\\
        WebCoSeg (BMVC 2017)~\cite{WebCoSeg} &  $56.4$ &  $56.9$\\
        DSNA (Arxiv 2018)~\cite{zhang2018decoupled}  & $58.2$  & $60.1$\\
        MDC (CVPR 2018)~\cite{wei2018revisiting} & $60.4$  & $60.8$\\
        MCOF (CVPR 2018)~\cite{wang2018weakly} & $60.3$  & $61.2$\\
        AffinityNet (CVPR 2018)~\cite{AffinityNet} &$61.7$ &$63.7$ \\
        Boostrap(CVPR 2018)~\cite{Bootstrap} & $63.0$ & $63.9$\\
        InstancesSalient(ECCV 2018)~\cite{InstancesSalient} & $64.5$ & $65.6$\\
        \textbf{Ours}   & $\textbf{61.9}$  & $\textbf{62.8}$\\
        \hline
        \bottomrule
    \end{tabular}
\end{table}

Fig.~\ref{fig:voc_performance} shows the importance of recursive refinement. The performance of all the networks improves as the training round increase and saturates after 3 or 4 training rounds. Fig.~\ref{fig:voc_performance} also shows that the performance of recursive training with different quality of masks: 1) by directly using the last round masks without any post-processing, and 2) by applying Grabcut on the last round masks. With GrabCut as the post-processing, the performance is better than directly using the masks from the last round due to the error propagation.  This phenomenon is consistent with the experiment in \cite{simple_does}.

\begin{table*}
    \footnotesize
    \caption{Per-class results on PASCAL VOC 2012 \emph{validation} and \emph{test} set.} 
    \setlength{\tabcolsep}{3pt}
    \begin{center}
        \begin{tabular}{ l | c c c c c c c c c c c c c c c c c c c c | c}
            \hline
            Method & aero & bike & bird & boat & bottle & bus & car & cat & chair & cow & table & dog & horse & mbike & person & plant & sheep & sofa & train & tv & mIoU\\
            \hline\hline
            Val & 77.1 & 25.9 & 75.3 & 59.8 & 62.3 & 80.2 & 73.9 & 79.7 & 16.9 & 70.7 & 32.5 & 73.1 & 72.2 & 67.8 & 69.2 & 45.0 & 72.6 & 42.6 & 72.3 & 41.9 & 61.9\\
            Test & 74.2 & 29.6 & 81.7 & 53.2 & 58.1 & 75.4 & 73.6 & 80.2 & 18.1 & 71.3 & 40.8 & 75.7 & 76.1 & 72.8 & 67.7 & 51.5 & 74.4 & 47.7 & 67.4 & 39.7 & 62.8\\ 
            \hline
        \end{tabular}
        \vspace{-5pt}
    \end{center}
    \label{tab:perclass}
    \vspace{-10pt}
\end{table*}

\begin{table*}[t]
\centering
\caption{Quantitative results on MIT Object Discovery dataset. Our proposed weakly supervised method achieves comparable performance as the several state-of-the-art supervised methods. ``*" indicates the method uses the human-labeled  pixel-level annotations.}
           \begin{tabular}{|c|c|c|c|c|c|c|c|c|}
               \hline
             {\bf Methods} & \multicolumn{4}{c|}{{\bf MIT dataset (subset)}} & \multicolumn{4}{c|}{{\bf MIT dataset (full)}} \\
\cline{2-9}
& {\bf Airplane} & {\bf Car} & {\bf Horse} & {\bf Average} & {\bf Airplane} & {\bf Car} & {\bf Horse} & {\bf Average} \\
\hline
{\bf \# Images} & 82 & 89  & 93 &N/A & 470 & 1208 & 810 &N/A \\
\hline                    
DiscrCoseg \cite{joulin2010discriminative} & 15.36 & 37.15  & 30.16  &27.56 & N/A & N/A & N/A & N/A\\
\hline
MCoSeg. \cite{joulin2012multi} & 11.72 & 35.15  & 29.53  &25.47 & N/A & N/A & N/A  &N/A \\
\hline
CoSegmentation \cite{kim2011distributed} & 7.9 &  0.04 & 6.43  &4.79  & N/A & N/A & N/A  &N/A \\
\hline
MITObject \cite{InternetObjectdataset} & 55.81 & 64.42 & 51.65  &57.26  & 55.62 & 63.35  & 53.88  &57.62 \\
\hline
EVK \cite{chen2014enriching} & 54.62 & 69.2  & 44.46   &56.09  & 60.87 & 62.74 &  60.23  &61.28 \\
\hline
ActiveSeg.~\cite{activesegmentation} & 58.65 & 66.47  & 53.57  &59.56   & 62.27 &  65.3 & 55.41  &60.99 \\
   \hline
MarkovChain \cite{MarkovChain} & 37.22 & 55.22  & 47.02  &46.49  & 41.52 & 54.34 & 49.67  &48.51 \\
   \hline
BooleanMap \cite{booleanmap} & 51.84 & 46.61  & 39.52  &45.99  & 54.09 & 47.38 & 44.12  &48.53 \\
     \hline
DeepMC ~\cite{DeepMC} & 41.75  & 59.16   & 39.34  &46.75  & 42.84 &    58.13 &    41.85  &47.61 \\  
     \hline
DeepSaliency~\cite{deepsaliency} & 69.11 & 83.48  & 57.61  &70.07  & 69.11 &    83.48    & 67.26  &73.28 \\ 
     \hline
MCG \cite{MCG} & 32.02 & 54.21  & 37.85  &40.27  & 35.32 & 52.98 & 40.44  &42.91 \\
    \hline 
DeepMask~\cite{DeepMask} & 71.81 & 67.01  & 58.80  &65.87  &    68.89 &    65.4 &    62.61  &65.63 \\
    \hline
            SalObj \cite{SalObj} & 53.91 & 58.03  & 47.42  &53.12  &  55.31 &    55.83    & 49.13  &53.42 \\
    \hline 
            UnsupervisedSeg \cite{forground_seg} & 61.37 & 70.52  & 55.09  &62.32  & N/A & N/A & N/A  &N/A \\
     \hline
        PixelObjectness* \cite{pixelobjectness} & 66.43 & 85.07  & 60.85  &70.78 & 66.18 & 84.80 & 64.90  &71.96 \\
 \hline
  {\bf Ours} & 64.92 &77.6  & 60.36  &{\bf 67.6} & 65.88 &77.07 & 65.82  &{\bf 69.9} \\
\hline
\end{tabular}

\label{tab:results_mit}
\end{table*}

\subsubsection{Comparison with others.} The performance of comparison on PASCAL VOC 2012 validation and test split is shown in Table~\ref{tab:val_comp}. Based on the level of annotations, these methods fall into two categories: object-level annotation-based methods and image-level annotation-based methods. The methods trained with accurate annotation of bounding box or spot for each object belong to the object-level annotation-based methods. Since the object-level annotations are more accurate and comprehensive than image-level annotations, these methods usually have better performance. For example, the methods trained with accurate bounding boxes have performance more than $60$\% IOU, while most of the methods with image-level labels have the performance less than $60$\%. Some image-level based methods implicitly use pixel-level supervision in their models such as \cite{AF_MCG}, \cite{TransferNet}, therefore their models can achieve relatively higher performance than those only using image-level labels.

Our method only uses the image-level annotations and achieves $61.9$\% on the validation split and $62.8$\% on the test split of Pascal VOC dataset. However, most of the image-level annotation-based methods are trained on PASCAL VOC dataset with accurate category label, while each image has multiple labels (see the examples of categories of Bird and Horse in Fig.~\ref{fig:refinement_visualize}).  Fan et al. \cite{InstancesSalient} proposed to use an instance-level salient object detector to produce salient instance which used as masks during training. Compared to the initial masks in our model, the instance-level saliency is a kind of stronger supervision. And it is reasonable that their model obtained the state-of-the-art performance ($65.6$\%) on the test split of VOC2012. Trained only with simple and inaccurate category label annotations, our model outperforms most of the image-level based methods. 

\begin{figure*}[!h]
\centering
\includegraphics[width=0.8\textwidth]{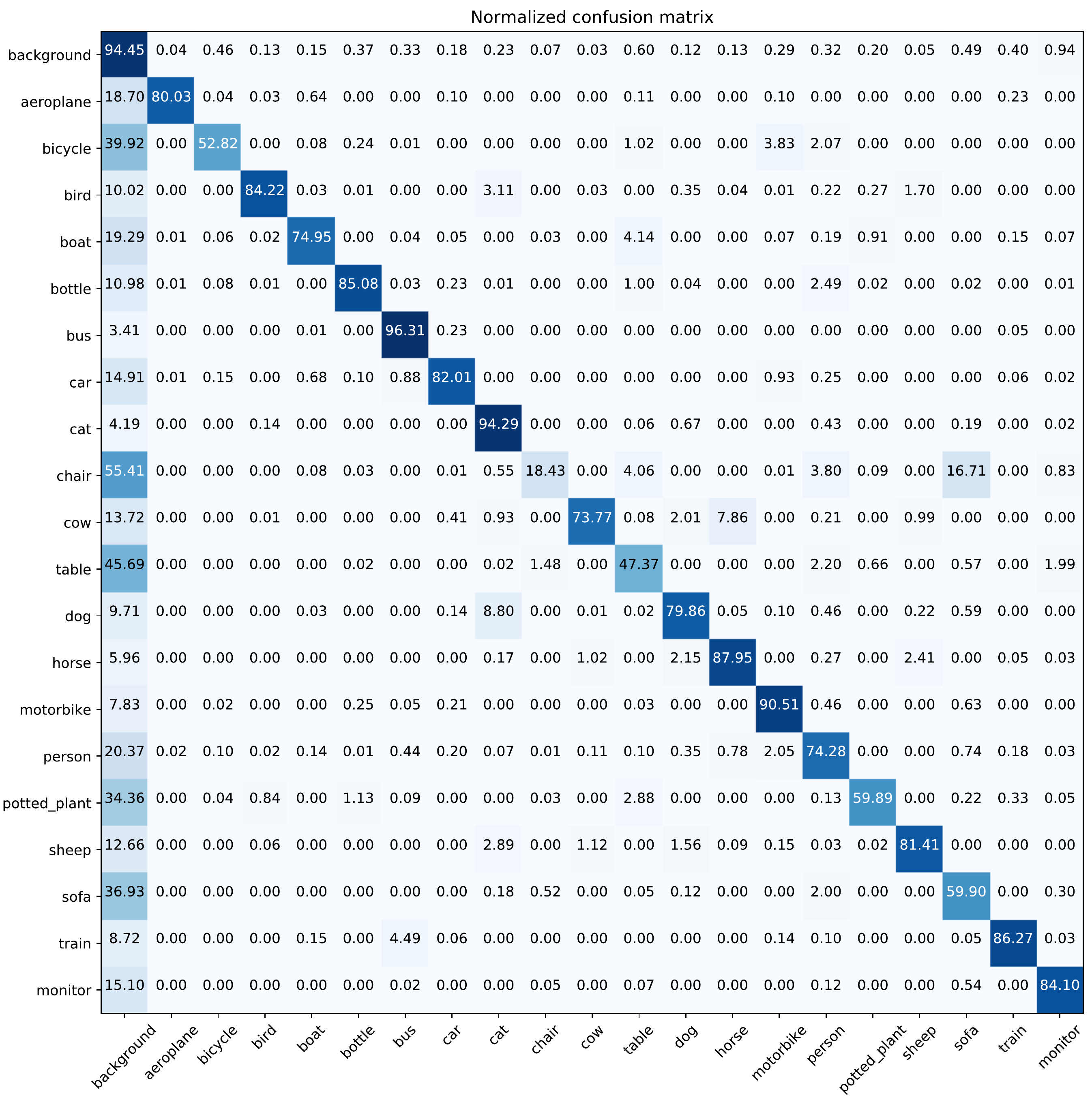}
\caption{The confusion matrix of our recursive semantic segmentation model. Our network can accurately classify the pixels of the most categories.}
\centering
\label{fig:confusion_matrix}
\end{figure*}

In addition to the final mean IOU result, we compute the per-class IOU as listed in Table~\ref{tab:perclass} and the confusion matrix of our model as shown in Fig.~\ref{fig:confusion_matrix}. Our model can accurately classify the pixels of most categories such as Aeroplane, Bird, Horse, Train, and Sheep, but has more errors in several categories including Bicycle, Chair, and Plant. For example, the pixel accuracy for Bird, Sheep, and Train categories is more than $80$\%. This demonstrates the effectiveness of the proposed method.

\subsection{Qualitative Results}

The recursive training the semantic segmentation network is a process of mask refinement. The coarse masks of training data are recursively refined by the semantic segmentation network. Some qualitative results of masks at different round are shown in Figs~\ref{fig:refinement_visualize}. Each image has only one category in the dataset. However, the trained semantic segmentation network can identify multi-categories in the training image. As shown in Fig.~\ref{fig:refinement_visualize}, the categories of Bird and Horse, with only one category label for each image, the final semantic segmentation masks can distinguish the birds and the chair, and the horse from persons.

\begin{figure}[!ht]
\centering
\includegraphics[width=0.5\textwidth]{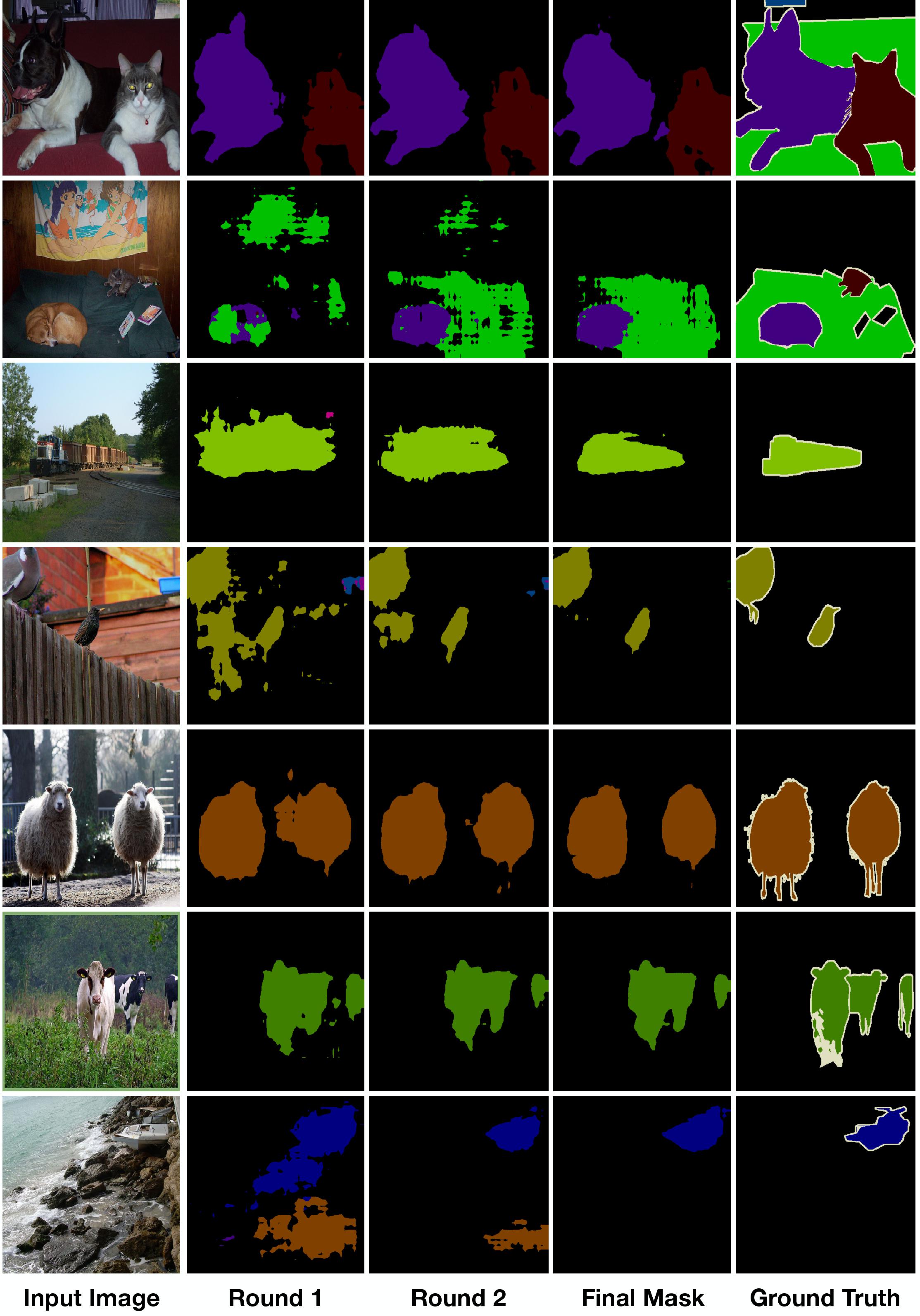}
\caption{More qualitative segmentation results by our semantic segmentation network. Different colors within one image mask represent different category objects.}
\label{fig:voc_visulaization}
\end{figure}

The qualitative results on PASCAL VOC dataset are shown in Fig.~\ref{fig:voc_visulaization}. Even though the coarse masks are very noisy, our network can predict the final pixel-wise semantic masks with only one category label available for each image. However, the results show that our network can handle images contain multi-category objects. There are objects belonging to multi-categories, but our trained model can distinguish the pixels of different categories (e.g. The first two rows in Fig.~\ref{fig:voc_visulaization}). 

\subsection{Foreground Segmentation Results}

The performance comparison on the test split of the MIT Object Discovery dataset \cite{InternetObjectdataset} is shown in Table~\ref{tab:results_mit}. Following \cite{pixelobjectness}, our proposed method is compared with $14$ existing state-of-the-art methods belonging to three categories: Joint Segmentation-based methods \cite{InternetObjectdataset}, \cite{joulin2010discriminative}, \cite{joulin2012multi}, \cite{kim2011distributed}, \cite{chen2014enriching}, \cite{activesegmentation}, saliency detection-based methods \cite{DeepMC}, \cite{deepsaliency}, \cite{MarkovChain}, \cite{booleanmap}, and object proposal-based methods \cite{MCG}, \cite{DeepMask}, \cite{SalObj}, \cite{pixelobjectness}. Trained with noisy masks, our proposed method achieves comparable performance as the state-of-the-art methods including the models trained with accurate human-labeled masks. 

Among all the methods, the performance of UnsupervisedSeg \cite{forground_seg} and PixelObjectness \cite{pixelobjectness} are most close to our method. The network in UnsupervisedSeg is trained with masks generated by an unsupervised method. Our model outperforms UnsupervisedSeg by $5.28$\% on the MIT Object Discovery dataset \cite{InternetObjectdataset}. The PixelObjectness is a supervised method in which a fully convolutional neural network is trained with accurate human-annotated masks to predict the masks. However, even trained with noisy masks, our proposed method achieves $69.9$\% which is only $2.06$\% lower than the supervised method.

\section{Conclusions}

We have proposed a novel coarse-to-fine semantic segmentation framework that can be trained from only image-level category labels and then iteratively refine the segmentation masks to pixel-wise level. The initial coarse mask is generated by a convolution neural network-based unsupervised foreground detection. Then a fully convolution neural network is recursively trained to continue to refine the masks. Finally, the final semantic segmentation mask is predicted by only use the simple image category label annotation. Our framework can handle images contains multiple categories of objects. With only trained on ImageNet, our framework achieves comparable performance on PASCAL VOC dataset as other image-level label-based state-of-the-arts of semantic segmentation and achieves comparable performance with the state-of-the-art supervised methods for the foreground object segmentation task.

\bibliographystyle{IEEEtran}
\bibliography{egbib}

\begin{thebibliography}{10}
\providecommand{\url}[1]{#1}
\csname url@samestyle\endcsname
\providecommand{\newblock}{\relax}
\providecommand{\bibinfo}[2]{#2}
\providecommand{\BIBentrySTDinterwordspacing}{\spaceskip=0pt\relax}
\providecommand{\BIBentryALTinterwordstretchfactor}{4}
\providecommand{\BIBentryALTinterwordspacing}{\spaceskip=\fontdimen2\font plus
\BIBentryALTinterwordstretchfactor\fontdimen3\font minus
  \fontdimen4\font\relax}
\providecommand{\BIBforeignlanguage}[2]{{%
\expandafter\ifx\csname l@#1\endcsname\relax
\typeout{** WARNING: IEEEtran.bst: No hyphenation pattern has been}%
\typeout{** loaded for the language `#1'. Using the pattern for}%
\typeout{** the default language instead.}%
\else
\language=\csname l@#1\endcsname
\fi
#2}}
\providecommand{\BIBdecl}{\relax}
\BIBdecl

\bibitem{FCN}
J.~Long, E.~Shelhamer, and T.~Darrell, ``Fully convolutional networks for
  semantic segmentation,'' in \emph{Proceedings of the IEEE conference on
  computer vision and pattern recognition}, 2015, pp. 3431--3440.

\bibitem{DeepLab}
L.-C. Chen, G.~Papandreou, I.~Kokkinos, K.~Murphy, and A.~L. Yuille, ``Deeplab:
  Semantic image segmentation with deep convolutional nets, atrous convolution,
  and fully connected crfs,'' \emph{IEEE transactions on pattern analysis and
  machine intelligence}, vol.~40, no.~4, pp. 834--848, 2018.

\bibitem{PSP}
H.~Zhao, J.~Shi, X.~Qi, X.~Wang, and J.~Jia, ``Pyramid scene parsing network,''
  in \emph{IEEE Conf. on Computer Vision and Pattern Recognition (CVPR)}, 2017,
  pp. 2881--2890.

\bibitem{SegNet}
V.~Badrinarayanan, A.~Kendall, and R.~Cipolla, ``Segnet: A deep convolutional
  encoder-decoder architecture for image segmentation,'' \emph{IEEE
  transactions on pattern analysis and machine intelligence}, vol.~39, no.~12,
  pp. 2481--2495, 2017.

\bibitem{VOC}
M.~Everingham, L.~Van~Gool, C.~K. Williams, J.~Winn, and A.~Zisserman, ``The
  pascal visual object classes (voc) challenge,'' \emph{International journal
  of computer vision}, vol.~88, no.~2, pp. 303--338, 2010.

\bibitem{Cityscape}
M.~Cordts, M.~Omran, S.~Ramos, T.~Rehfeld, M.~Enzweiler, R.~Benenson,
  U.~Franke, S.~Roth, and B.~Schiele, ``The cityscapes dataset for semantic
  urban scene understanding,'' in \emph{Proceedings of the IEEE conference on
  computer vision and pattern recognition}, 2016, pp. 3213--3223.

\bibitem{CamVid}
G.~J. Brostow, J.~Fauqueur, and R.~Cipolla, ``Semantic object classes in video:
  A high-definition ground truth database,'' \emph{Pattern Recognition
  Letters}, vol.~30, no.~2, pp. 88--97, 2009.

\bibitem{ADE20K}
B.~Zhou, H.~Zhao, X.~Puig, S.~Fidler, A.~Barriuso, and A.~Torralba, ``Scene
  parsing through ade20k dataset,'' in \emph{Proceedings of the IEEE Conference
  on Computer Vision and Pattern Recognition}, 2017.

\bibitem{ImageNet}
A.~Krizhevsky, I.~Sutskever, and G.~E. Hinton, ``Imagenet classification with
  deep convolutional neural networks,'' in \emph{Advances in neural information
  processing systems}, 2012, pp. 1097--1105.

\bibitem{YoutubeBB}
E.~Real, J.~Shlens, S.~Mazzocchi, X.~Pan, and V.~Vanhoucke,
  ``Youtube-boundingboxes: A large high-precision human-annotated data set for
  object detection in video,'' in \emph{2017 IEEE Conference on Computer Vision
  and Pattern Recognition (CVPR)}.\hskip 1em plus 0.5em minus 0.4em\relax IEEE,
  2017, pp. 7464--7473.

\bibitem{semi_gan}
N.~Souly, C.~Spampinato, and M.~Shah, ``Semi and weakly supervised semantic
  segmentation using generative adversarial network,'' \emph{arXiv preprint
  arXiv:1703.09695}, 2017.

\bibitem{web_videos}
S.~Hong, D.~Yeo, S.~Kwak, H.~Lee, and B.~Han, ``Weakly supervised semantic
  segmentation using web-crawled videos,'' \emph{arXiv preprint
  arXiv:1701.00352}, 2017.

\bibitem{simple_does}
A.~Khoreva, R.~Benenson, J.~Hosang, M.~Hein, and B.~Schiele, ``Simple does it:
  Weakly supervised instance and semantic segmentation,'' in \emph{Proc. CVPR},
  2017.

\bibitem{region_mining}
Y.~Wei, J.~Feng, X.~Liang, M.-M. Cheng, Y.~Zhao, and S.~Yan, ``Object region
  mining with adversarial erasing: A simple classification to semantic
  segmentation approach,'' in \emph{IEEE CVPR}, 2017.

\bibitem{scribble}
D.~Lin, J.~Dai, J.~Jia, K.~He, and J.~Sun, ``Scribblesup: Scribble-supervised
  convolutional networks for semantic segmentation,'' in \emph{Proceedings of
  the IEEE Conference on Computer Vision and Pattern Recognition}, 2016, pp.
  3159--3167.

\bibitem{BoxSup}
J.~Dai, K.~He, and J.~Sun, ``Boxsup: Exploiting bounding boxes to supervise
  convolutional networks for semantic segmentation,'' in \emph{Proceedings of
  the IEEE International Conference on Computer Vision}, 2015, pp. 1635--1643.

\bibitem{WSSL}
G.~Papandreou, L.-C. Chen, K.~Murphy, and A.~L. Yuille, ``Weakly-and
  semi-supervised learning of a dcnn for semantic image segmentation,''
  \emph{arXiv preprint arXiv:1502.02734}, 2015.

\bibitem{one_point}
A.~Bearman, O.~Russakovsky, V.~Ferrari, and L.~Fei-Fei, ``What’s the point:
  Semantic segmentation with point supervision,'' in \emph{European Conference
  on Computer Vision}.\hskip 1em plus 0.5em minus 0.4em\relax Springer, 2016,
  pp. 549--565.

\bibitem{AF_MCG}
X.~Qi, Z.~Liu, J.~Shi, H.~Zhao, and J.~Jia, ``Augmented feedback in semantic
  segmentation under image level supervision,'' in \emph{European Conference on
  Computer Vision}.\hskip 1em plus 0.5em minus 0.4em\relax Springer, 2016, pp.
  90--105.

\bibitem{forground_seg}
I.~Croitoru, S.-V. Bogolin, and M.~Leordeanu, ``Unsupervised learning from
  video to detect foreground objects in single images,'' in \emph{The IEEE
  International Conference on Computer Vision (ICCV)}, Oct 2017.

\bibitem{TransferNet}
S.~Hong, J.~Oh, H.~Lee, and B.~Han, ``Learning transferrable knowledge for
  semantic segmentation with deep convolutional neural network,'' in
  \emph{Proceedings of the IEEE Conference on Computer Vision and Pattern
  Recognition}, 2016, pp. 3204--3212.

\bibitem{SEC}
A.~Kolesnikov and C.~H. Lampert, ``Seed, expand and constrain: Three principles
  for weakly-supervised image segmentation,'' in \emph{European Conference on
  Computer Vision}.\hskip 1em plus 0.5em minus 0.4em\relax Springer, 2016, pp.
  695--711.

\bibitem{wei2018revisiting}
Y.~Wei, H.~Xiao, H.~Shi, Z.~Jie, J.~Feng, and T.~S. Huang, ``Revisiting dilated
  convolution: A simple approach for weakly-and semi-supervised semantic
  segmentation,'' in \emph{Proceedings of the IEEE Conference on Computer
  Vision and Pattern Recognition}, 2018, pp. 7268--7277.

\bibitem{zhang2018decoupled}
T.~Zhang, G.~Lin, J.~Cai, T.~Shen, C.~Shen, and A.~C. Kot, ``Decoupled spatial
  neural attention for weakly supervised semantic segmentation,'' \emph{arXiv
  preprint arXiv:1803.02563}, 2018.

\bibitem{InternetObjectdataset}
M.~Rubinstein, A.~Joulin, J.~Kopf, and C.~Liu, ``Unsupervised joint object
  discovery and segmentation in internet images,'' in \emph{Proceedings of the
  IEEE conference on computer vision and pattern recognition}, 2013, pp.
  1939--1946.

\bibitem{joulin2010discriminative}
A.~Joulin, F.~Bach, and J.~Ponce, ``Discriminative clustering for image
  co-segmentation,'' in \emph{Computer Vision and Pattern Recognition (CVPR),
  2010 IEEE Conference on}.\hskip 1em plus 0.5em minus 0.4em\relax IEEE, 2010,
  pp. 1943--1950.

\bibitem{joulin2012multi}
------, ``Multi-class cosegmentation,'' in \emph{Computer Vision and Pattern
  Recognition (CVPR), 2012 IEEE Conference on}.\hskip 1em plus 0.5em minus
  0.4em\relax IEEE, 2012, pp. 542--549.

\bibitem{kim2011distributed}
G.~Kim, E.~P. Xing, L.~Fei-Fei, and T.~Kanade, ``Distributed cosegmentation via
  submodular optimization on anisotropic diffusion,'' in \emph{2011
  International Conference on Computer Vision}.\hskip 1em plus 0.5em minus
  0.4em\relax IEEE, 2011, pp. 169--176.

\bibitem{chen2014enriching}
X.~Chen, A.~Shrivastava, and A.~Gupta, ``Enriching visual knowledge bases via
  object discovery and segmentation,'' in \emph{Proceedings of the IEEE
  conference on computer vision and pattern recognition}, 2014, pp. 2027--2034.

\bibitem{activesegmentation}
S.~Dutt~Jain and K.~Grauman, ``Active image segmentation propagation,'' in
  \emph{Proceedings of the IEEE Conference on Computer Vision and Pattern
  Recognition}, 2016, pp. 2864--2873.

\bibitem{DeepMC}
R.~Zhao, W.~Ouyang, H.~Li, and X.~Wang, ``Saliency detection by multi-context
  deep learning,'' in \emph{Proceedings of the IEEE Conference on Computer
  Vision and Pattern Recognition}, 2015, pp. 1265--1274.

\bibitem{deepsaliency}
X.~Li, L.~Zhao, L.~Wei, M.-H. Yang, F.~Wu, Y.~Zhuang, H.~Ling, and J.~Wang,
  ``Deepsaliency: Multi-task deep neural network model for salient object
  detection,'' \emph{IEEE Transactions on Image Processing}, vol.~25, no.~8,
  pp. 3919--3930, 2016.

\bibitem{MarkovChain}
B.~Jiang, L.~Zhang, H.~Lu, C.~Yang, and M.-H. Yang, ``Saliency detection via
  absorbing markov chain,'' in \emph{The IEEE International Conference on
  Computer Vision (ICCV)}, December 2013.

\bibitem{booleanmap}
J.~Zhang and S.~Sclaroff, ``Saliency detection: A boolean map approach,'' in
  \emph{Proceedings of the IEEE international conference on computer vision},
  2013, pp. 153--160.

\bibitem{MCG}
P.~Arbel{\'a}ez, J.~Pont-Tuset, J.~T. Barron, F.~Marques, and J.~Malik,
  ``Multiscale combinatorial grouping,'' in \emph{Proceedings of the IEEE
  conference on computer vision and pattern recognition}, 2014, pp. 328--335.

\bibitem{DeepMask}
P.~O. Pinheiro, R.~Collobert, and P.~Doll{\'a}r, ``Learning to segment object
  candidates,'' in \emph{Advances in Neural Information Processing Systems},
  2015, pp. 1990--1998.

\bibitem{SalObj}
Y.~Li, X.~Hou, C.~Koch, J.~M. Rehg, and A.~L. Yuille, ``The secrets of salient
  object segmentation,'' in \emph{Proceedings of the IEEE Conference on
  Computer Vision and Pattern Recognition}, 2014, pp. 280--287.

\bibitem{pixelobjectness}
S.~D. Jain, B.~Xiong, and K.~Grauman, ``Pixel objectness,'' \emph{arXiv
  preprint arXiv:1701.05349}, 2017.

\bibitem{VideoPCA}
O.~Stretcu and M.~Leordeanu, ``Multiple frames matching for object discovery in
  video.'' in \emph{BMVC}, vol.~1, no.~2, 2015.

\bibitem{NLC}
A.~Faktor and M.~Irani, ``Video segmentation by non-local consensus voting.''
  in \emph{BMVC}, vol.~2, no.~7, 2014.

\bibitem{VideoPCA_new}
E.~Haller and M.~Leordeanu, ``Unsupervised object segmentation in video by
  efficient selection of highly probable positive features,'' \emph{arXiv
  preprint arXiv:1704.05674}, 2017.

\bibitem{GrabCut}
C.~Rother, V.~Kolmogorov, and A.~Blake, ``Grabcut: Interactive foreground
  extraction using iterated graph cuts,'' in \emph{ACM transactions on graphics
  (TOG)}, vol.~23, no.~3.\hskip 1em plus 0.5em minus 0.4em\relax ACM, 2004, pp.
  309--314.

\bibitem{FPN}
T.-Y. Lin, P.~Doll{\'a}r, R.~B. Girshick, K.~He, B.~Hariharan, and S.~J.
  Belongie, ``Feature pyramid networks for object detection.'' in \emph{CVPR},
  vol.~1, no.~2, 2017, p.~4.

\bibitem{BFBP}
F.~Saleh, M.~S. Aliakbarian, M.~Salzmann, L.~Petersson, S.~Gould, and J.~M.
  Alvarez, ``Built-in foreground/background prior for weakly-supervised
  semantic segmentation,'' in \emph{European Conference on Computer
  Vision}.\hskip 1em plus 0.5em minus 0.4em\relax Springer, 2016, pp. 413--432.

\bibitem{MILSEG}
P.~O. Pinheiro and R.~Collobert, ``From image-level to pixel-level labeling
  with convolutional networks,'' in \emph{Proceedings of the IEEE Conference on
  Computer Vision and Pattern Recognition}, 2015, pp. 1713--1721.

\bibitem{MIL_FCN}
D.~Pathak, E.~Shelhamer, J.~Long, and T.~Darrell, ``Fully convolutional
  multi-class multiple instance learning,'' in \emph{ICLR Workshop}, 2015.

\bibitem{CCCN}
D.~Pathak, P.~Krahenbuhl, and T.~Darrell, ``Constrained convolutional neural
  networks for weakly supervised segmentation,'' in \emph{Proceedings of the
  IEEE international conference on computer vision}, 2015, pp. 1796--1804.

\bibitem{DCSM}
W.~Shimoda and K.~Yanai, ``Distinct class-specific saliency maps for weakly
  supervised semantic segmentation,'' in \emph{European Conference on Computer
  Vision}.\hskip 1em plus 0.5em minus 0.4em\relax Springer, 2016, pp. 218--234.

\bibitem{STC}
Y.~Wei, X.~Liang, Y.~Chen, X.~Shen, M.-M. Cheng, J.~Feng, Y.~Zhao, and S.~Yan,
  ``Stc: A simple to complex framework for weakly-supervised semantic
  segmentation,'' \emph{IEEE transactions on pattern analysis and machine
  intelligence}, vol.~39, no.~11, pp. 2314--2320, 2017.

\bibitem{WeblySeg}
B.~Jin, M.~V.~O. Segovia, and S.~Süsstrunk, ``Webly supervised semantic
  segmentation,'' in \emph{2017 IEEE Conference on Computer Vision and Patten
  Recognition (CVPR)}, 2017.

\bibitem{WebCoSeg}
T.~Shen, G.~Lin, L.~Liu, C.~Shen, and I.~Reid, ``Weakly supervised semantic
  segmentation based on web image co-segmentation,'' in \emph{BMVC}, 2017.

\bibitem{wang2018weakly}
X.~Wang, S.~You, X.~Li, and H.~Ma, ``Weakly-supervised semantic segmentation by
  iteratively mining common object features,'' in \emph{Proceedings of the IEEE
  Conference on Computer Vision and Pattern Recognition}, 2018, pp. 1354--1362.

\bibitem{AffinityNet}
J.~Ahn and S.~Kwak, ``Learning pixel-level semantic affinity with image-level
  supervision for weakly supervised semantic segmentation,'' in \emph{The IEEE
  Conference on Computer Vision and Pattern Recognition (CVPR)}, June 2018.

\bibitem{Bootstrap}
T.~Shen, G.~Lin, C.~Shen, and I.~Reid, ``Bootstrapping the performance of webly
  supervised semantic segmentation,'' in \emph{IEEE Conference on Computer
  Vision and Pattern Recognition (CVPR'18)}, 2018.

\bibitem{InstancesSalient}
S.-M. Hu, ``Associating inter-image salient instances for weakly supervised
  semantic segmentation,'' 2018.

\end{thebibliography}
\end{document}